%% file: main.tex
 \documentclass[pmlr,twocolumn,10pt]{jmlr} 

\usepackage{booktabs}
\usepackage{siunitx}

\theorembodyfont{\upshape}
\theoremheaderfont{\scshape}
\theorempostheader{:}
\theoremsep{\newline}

\jmlrvolume{259}
\jmlryear{2024}
\jmlrsubmitted{LEAVE UNSET}
\jmlrpublished{LEAVE UNSET}
\jmlrworkshop{Machine Learning for Health (ML4H) 2024} 

 \title[Labrador]{Labrador: Exploring the limits of masked language modeling for laboratory data}

\newcommand{\labrador}{{\sc Labrador}\xspace}
\usepackage{threeparttable}
\usepackage{makecell}

\author{%
\Name{David R. Bellamy\nametag{\thanks{Corresponding author.}}}$^{1,2}$ \Email{bellamyrd@gmail.com}\\
\Name{Bhawesh Kumar}$^{2}$ \Email{bhawesh\_kumar@hsph.harvard.edu}\\
\Name{Cindy Wang}$^{3}$ \Email{cindywang@college.harvard.edu}\\
\Name{Andrew Beam}$^{1,2}$ \Email{andrew\_beam@hms.harvard.edu}\\
\addr $^{1}$ Harvard Epidemiology Department \\
\addr $^{2}$ Harvard Biostatistics Department \\
\addr $^{3}$ Harvard Statistics Department
}

\begin{document}

\maketitle

\begin{abstract}
\input{sections/abstract}
\end{abstract}

\paragraph*{Data and Code Availability}
\input{sections/code_and_data}

\section{Introduction}
\input{sections/introduction}

\begin{figure*}[tb]
\centering
\includegraphics[width=0.65\linewidth]{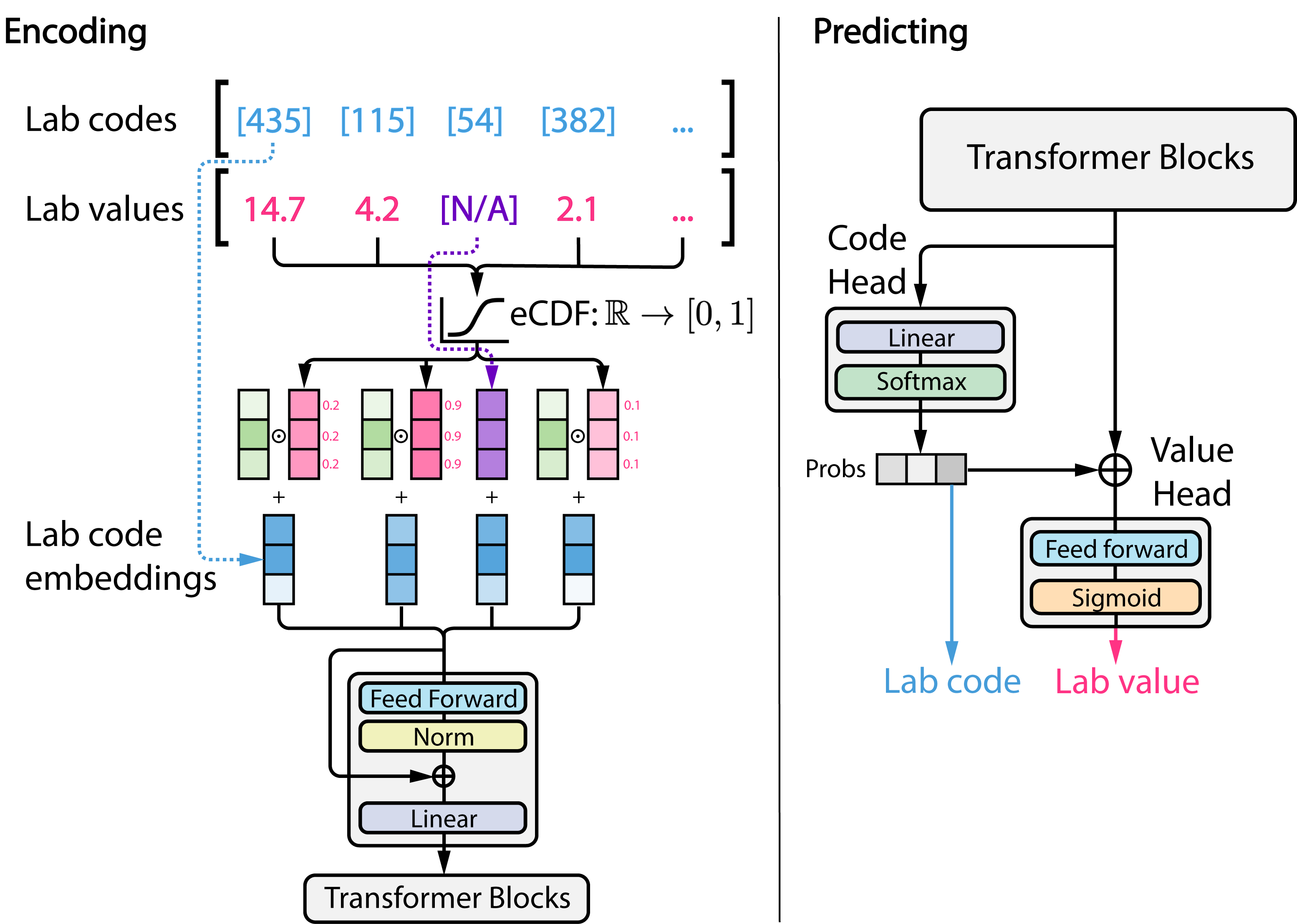}
\caption{\labrador model architecture.}
\label{fig:fig1}
\end{figure*}

\section{Related Work}
\input{sections/background}

\section{Methods}
\label{sec:methods}
\input{sections/methods}

\section{Results}
\label{sec:results}
\input{sections/results}

\section{Discussion}
\label{sec:discussion}
\input{sections/discussion}

\section{Conclusion}
\input{sections/conclusion}

\bibliography{references}

\appendix
\input{sections/appendix}

\end{document}

%% file: sections/abstract.tex
In this work we introduce \labrador, a pre-trained Transformer model for laboratory data. \labrador and BERT were pre-trained on a corpus of 100 million lab test results from electronic health records (EHRs) and evaluated on various downstream outcome prediction tasks. Both models demonstrate mastery of the pre-training task but neither consistently outperform XGBoost on downstream supervised  tasks. Our ablation studies reveal that transfer learning shows limited effectiveness for BERT and achieves marginal success with \labrador. We explore the reasons for the failure of transfer learning and suggest that the data generating process underlying each patient cannot be characterized sufficiently using labs alone, among other factors. We encourage future work to focus on joint modeling of multiple EHR data categories and to include tree-based baselines in their evaluations.

%% file: sections/code_and_data.tex
The MIMIC-IV database can be accessed via PhysioNet after completing a short online training. The lab data that
was used to pre-train \labrador and BERT come from the labevents
table within MIMIC-IV and our pre-processing code can be found in our \href{https://github.com/DavidBellamy/labrador}{codebase}. The COVID-19, cancer diagnosis and alcohol consumption fine-tuning datasets are public and therefore we provide our pre-processed versions of these datasets directly in our codebase. 

The codebase also contains all of the code
necessary to replicate the pre-training data pre-processing, model pre-training,
fine-tuning and other evaluations. We cannot directly share the
sepsis dataset due to MIMIC-IV's data sharing policy, however with access to MIMIC-IV the steps described in Section \ref{sepsis-cohort} recreate our evaluation dataset. Finally, the weights for the pre-trained \labrador and BERT models can be downloaded from our HuggingFace repository \href{https://huggingface.co/Drbellamy/labrador}{here}.

%% file: sections/introduction.tex
In recent years, self-supervised pre-training of masked language models
(MLMs) (see Appendix \ref{appendix_background} for background) has demonstrated remarkable
success across a wide range of machine learning problems and has led to significant downstream improvements across diverse tasks in
natural language
processing \citep{liu2019roberta, devlin-etal-2019-bert, raffel2020exploring}.
There is considerable excitement surrounding the potential of large
pre-trained MLMs to achieve similar success in medical applications. For
instance, existing applications of MLMs in medicine have already yielded
promising results in tasks related to medical text
understanding \citep{Lee2020-vq, alsentzer-etal-2019-publicly, huang2019clinicalbert, yang-etal-2019-enhancing-pre, beltagy-etal-2019-scibert}.
Despite the success of self-supervised models in certain areas of
biomedicine \citep{palepu2022tier, Beam2023-qu, singhal2023large, Moor2023-ue, Tiu2022-yt, Ingraham2022-sr, Watson2023-do, Shay2023-vn},
there has been no previous work on  pre-trained models for laboratory
measurements. Laboratory data is abundant, routinely collected, less
biased compared to other types of data in electronic health records
(EHRs) like billing
codes \citep{Beam2021-tw},
and directly measure a patient's physiological state, offering a
valuable opportunity for creating a medical foundation
model.

However, there is a large body of
evidence showing that deep learning is consistently outperformed on
so-called ``tabular'' data prediction tasks by traditional machine learning techniques
like random forests, XGBoost, and even simple regression
models \citep{Bellamy2020-qr, finlayson2023machine, Sharma2013-ph}.
The reasons for this are only partially understood, but previous
work \citep{grinsztajn2022tree} has
suggested that this phenomenon may be caused by a rotational invariance
in deep learning models that is harmful for tabular data. More broadly, the
success of deep learning is thought to be largely due to inductive
biases that can be leveraged for images, text, and graphs. These
inductive biases are absent or only weakly present in tabular data.
Conversely, tree-based methods are scale invariant and robust to
uninformative features.

In this work, we introduce \labrador, a novel continuous Transformer architecture, which models permutation-invariant data consisting
of (integer, float) tuples (Figure \ref{fig:fig1}). We pre-trained \labrador alongside a standard
implementation of BERT on 100 million lab test results
from over 260,000 patients using an MLM objective. We evaluated both models
on several downstream outcome prediction tasks and validated the success of pre-training with a set of intrinsic evaluations.

We discuss the design and implementation of \labrador and BERT, their training process, and their evaluation in Sections \ref{sec:methods} and \ref{sec:results}, respectively. We conclude in Section \ref{sec:discussion} with the implications of our findings, shedding light on the limitations and
areas for improvement in the application of pre-trained MLMs to
laboratory data and more generally, the creation of foundation models for numeric EHR data.

%% file: sections/background.tex
In previous work, the Transformer
architecture \citep{Vaswani2017-hn}
has been adapted to handle continuous inputs alongside discrete inputs (i.e. tokens). For example, \cite{gorishniy2021revisiting} introduce the Feature-Tokenizer (FT) Transformer, which is a similar
architecture to \labrador. They benchmark FT
Transformer against gradient boosted decision trees and Random Forests
on several supervised learning tasks (without pre-training) and conclude
that there is no universally superior solution. Later
work by
these authors \citep{gorishniy2022embeddings} introduced an additional variant of continuous Transformer
that uses a piecewise linear encoding system for continuous values.
These authors argue that the design of embeddings for numerical features
is an important consideration in adapting Transformer models to
continuous data, a notion that we support strongly. Our work may be
viewed as an extension of the FT Transformer to the setting of MLM-based
pre-training.

\cite{rossi2019evaluation} learned embeddings of laboratory data using
Word2Vec \citep{mikolov2013efficient}
and
GloVe \citep{Pennington2014-ut} by encoding each lab test result as the concatenation of an octal indicator of abnormality and the test's
LOINC code\footnote{\url{https://loinc.org/get-started/what-loinc-is/}}. The ablations showed improved mortality prediction when the octal indicator was included in the embeddings despite this being only a discrete representation of the underlying continuous lab test result.

\cite{pmlr-v206-hegselmann23a} showed that pre-trained large language models (LLMs) can directly perform prediction tasks involving numeric features by
serializing the features and a description of the task into a natural
language string -- an approach the authors called TabLLM. They found
that this approach could outperform strong tree-based baselines in the
very-few-shot setting (0 to 8 labeled examples). However, \cite{dinh2022lift}
could not reproduce this result with GPT-3 and instead found that a
fine-tuned GPT-3 model performed worse than logistic regression for up
to 250 training examples. Hegselmann et al. also found that the
downstream performance is highly task dependent and that performance on
medical prediction tasks was notably worse than others in a more general
knowledge domain. However, regardless of the domain, XGBoost outperforms
TabLLM on average across all tasks.

TabTransformer \citep{huang2020tabtransformer}
is another architecture that can model categorical and continuous
numeric features, however only categorical features pass through the attention layer. Numeric features are passed directly to the
prediction head where they are concatenated to the contextualized embeddings for
the categorical features. TabTransformer is distinct from our
architecture, which handles data that consist of (integer, float)
samples and performs attention jointly over these two types of data. To the best of our knowledge, the recently introduced xVal encoding scheme \citep{golkar2023xval} is the only existing work that adapts the Transformer architecture to jointly model these two types of data during pre-training. Their multi-task MLM pre-training objective and model architecture are similar to \labrador so we designed Figure \ref{fig:fig1} to facilitate their comparison. However, the authors evaluate xVal on a distinct set of tasks (learning arithmetic, temperature forecasting and predicting planetary orbits) and do not include a baseline comparison with gradient boosted decision trees.

Recent work by \citet{zhu2024larger} evaluated 14 language models alongside conventional clinical predictive models and found that even with well-designed prompts, the best LLMs' zero-shot performance does not consistently beat traditional machine learning models when there are 100 or more examples to train on.

%% file: sections/methods.tex
In this section, we describe the \labrador and BERT model architectures and pre-training. For further experimental details, see Appendix \ref{appendix_further_exp_details}.

BERT is primarily distinguished from other Transformers by its learning objective rather than its model architecture. BERT's architecture is classified as encoder-only and consists of an embedding layer, N transformer blocks with the canonical layers (MultiHeadAttention, Dropout, LayerNorm, Feedforward, Dropout, LayerNorm), and a prediction head.

The primary contribution of the original work was the approach of masked language modeling (MLM). The key advantage of MLM is that it does not rely on the input data having a natural sequential ordering, unlike autoregressive language modeling. The lab data in our study does not have a natural sequential ordering so we had to use an MLM-based approach. We considered ways to add positional embeddings to the lab data such that they would have a sequential ordering (even if only artificially), but we concluded that it is impossible to find a natural ordering because batches of lab tests are requested at the same moment in time by physicians. So there is no way to order labs sequentially besides alphabetically or some other arbitrary order. We did not think that adding such arbitrary ordering would be helpful. 

Besides encoder-only transformer architectures, GPT-style models have decoder-only architectures and older seq2seq transformer models (e.g. t5) have encoder-decoder architectures. Both decoder-only and encoder-decoder architectures are useful for generating sequences like text. In other words, these architectures were designed for seq2seq tasks. But since lab data do not have a sequential ordering, we cannot pose it as a seq2seq task so these transformer architectures are not applicable to our setting. This is why we chose a BERT-like (i.e. encoder-only) MLM-based approach in this work.

\subsection{Labrador architecture}\label{labrador-continuous-transformer-architecture}

\labrador consists of two embedding modules (categorical and continuous), N transformer blocks, and two prediction heads (categorical and continuous). The categorical embedding layer uses the standard integer lookup method, whereas the continuous embedding layer involves a position-wise linear projection. A pseudo-code implementation is in Section \ref{pseudocode} of the Appendix. 

Each Transformer block consists of the canonical layer ordering: MultiHeadAttention, Dropout, LayerNorm, Feedforward, Dropout, LayerNorm. However, we use a distinct \verb|key_dim| size for the multi-head attention layers. We set this hyperparameter equal to \labrador's embedding dimension, whereas it is commonly set to be equal to \verb|d_model // num_heads| where \verb|d_model| is the model's embedding size and \verb|num_heads| is the number of attention heads.

The categorical prediction head consists of a ReLU-activated
position-wise dense layer followed by a softmax layer over the 529 token
vocabulary. A pseudo-code implementation of the continuous prediction head is also in Section \ref{pseudocode} of the Appendix. The output of this layer is simply a float
on the interval $[0, 1]$, since all input values are standardized
to this interval using the empirical cumulative distribution function
(see Section \ref{preprocessing-of-the-pre-training-lab-data}). In our evaluations, we use 10 transformer blocks, an embedding dimension of 1024, 4 attention heads, and a feedforward dimension of 1024, resulting in 196,642,645 parameters.

\subsection{BERT architecture}\label{bert-architecture}

BERT is implemented using HuggingFace's \verb|TFBertForMaskedLM|. To eliminate positional embeddings, we pass the zero vector in place of the \verb|position_ids| key.

The model configuration (HuggingFace's \verb|BERTConfig|) that is used in our experiments has a vocabulary size of 4251, a hidden size of 1024, 10 hidden layers, 4 attention heads and an intermediate (i.e. feedforward) size of 1024 with a ReLU activation.

All other parameters use their default \verb|BERTConfig| value. Most notably, the dropout probabilities are set to 0.1 throughout the model, which is consistent with \labrador and common practice.

This model shares the same feedforward layer size
(\verb|intermediate_size|), number of attention heads and transformer
blocks (\verb|num_attention_heads| and \verb|num_hidden_layers|), dropout
rate, and embedding size (\verb|hidden_size|) with \labrador. Therefore, this BERT model has the same backbone as \labrador and differs only in
its embedding module, prediction head and the \verb|key_dim| in the multi-head attention layers. 

This BERT instance has 68,522,139 parameters, which is less than \labrador due to their difference in \verb|key_dim| size. To ensure that this difference in parameter count does not impact our conclusions, we performed all experiments with a parameter-matched BERT that uses the same \verb|key_dim| as \labrador (Appendix \ref{appendix_parameter_matched}). The results from these experiments are indistinguishable from the results in Section \ref{sec:results}.

\subsection{Preprocessing of pre-training lab data}\label{preprocessing-of-the-pre-training-lab-data}

We pre-processed the lab data obtained from MIMIC
(version IV) before MLM pre-training. First, we removed all lab codes (i.e. ``itemid'') from the labevents data table that had 500 or fewer occurrences to ensure that there
was a sufficient number of samples for BERT and \labrador to learn about
each lab code. This results in 529 unique lab codes. We split patients into training, validation and test sets. 70\%
($\approx 229,000$) of the patients were selected uniformly at
random for the training split, whereas 10\% and 20\% formed the
validation and test splits, respectively. 

For each lab code, the
empirical cumulative distribution function (eCDF) is computed using the
training split. Normally, mapping values of a random variable to their probabilities under an eCDF requires the entire dataset. However, in practice, there are far fewer unique lab values in the dataset. Therefore, we can represent each eCDF using only the unique lab values for each lab code as well as the probabilities that they map to on their corresponding eCDF. This compression of the eCDF is lossless, which means that it perfectly represents the eCDF obtained from the full training split. These compressed
eCDFs are used in all downstream pre-training and evaluation steps and can be found in the \verb|data/| directory of our codebase.

To tokenize the lab data for \labrador,
lab values are mapped to the interval $[0, 1]$ using their eCDFs whereas
lab codes are mapped to their integer frequency ranking in the training split, indexed at 1. For example, hematocrit, the most frequently ordered lab test, receives the integer token 1. We also include a mask
token (the integer 530) and a special null token (the integer 531). The null token is used when a lab test is recorded but it has no associated lab value. Approximately 10\% of lab tests in MIMIC-IV have no associated lab value. Although we used frequency ranking to assign integer tokens to lab codes, this ordering has no consequence so even an arbitrary ordering of the lab codes could be used. 

For BERT, we construct an integer-only token vocabulary by assigning a unique integer to each decile of
the eCDF for each lab code, sorted in descending order of test
frequency. For example, tokens 1 through 10 represent the 10 deciles of the hematocrit lab value distribution because it is the lab test with the highest frequency ranking. We also include an
eleventh token that represents a missing value for
hematocrit. The second most frequent test was creatinine and therefore tokens 12 through 21 represent the 10
deciles of its lab value distribution. Once again, token 22 stands
for a missing value of creatinine. This produces a vocabulary size of
4250 tokens plus one additional token that represents the
\verb|<MASK>| indicator. Although there are 529 unique
lab codes, there are not $529 \times 11 = 5819$ tokens in BERT's vocabulary
because 157 of the lab codes have no numeric values and, instead, are
interpreted as binary presence/absence indicators. For example, the
toxicology screens for opiates, cocaine, amphetamines and barbiturates
are entered as positive/negative test results.

Finally, the labs for each patient are divided into order sets. A bag is defined as all labs ordered at the same time for a specific patient in the MIMIC database, also called an order set. Bags with fewer than 3 labs are removed from
the training, validation and test splits as we consider these bags to be of insufficient size for learning  embeddings.
Order sets are converted to an appropriate input structure for
\labrador and BERT separately. For \labrador, each bag is structured as a
dictionary with keys \verb|categorical_input| (integer) and
\verb|continuous_input| (float), whereas BERT's input dictionary has keys \verb|input_ids| (all integers) and
\verb|position_ids| (the zero vector). Bags of different lengths are padded to the
maximum size in the current batch. In order to minimize step time during pre-training, we performed random masking of one element per bag
in advance and sharded these bags across approximately 220 TensorFlow TFRecord files.

\subsection{Transformer pre-training}\label{transformer-pre-training}

Both \labrador and BERT were pre-trained using a single 40GB A100 GPU for
500,000 steps (80 hours) and 1.5M steps (237 hours), respectively. Our
stopping criterion for each model was to observe that the validation
loss was sufficiently converged. We also wanted to ensure that BERT's
degree of convergence was greater than or equal to \labrador's in order
to avoid falsely representing the capabilities of BERT in downstream
evaluations. Both models were pre-trained with a dropout rate of 0.1, learning rate of 1e-5, a batch size of 256, and an embedding size of 1024. 

We used Adam optimization for stochastic gradient descent. Model
checkpoints were saved every 14,000 training steps. BERT optimized a
standard categorical cross-entropy loss for measuring its accuracy at
predicting masked tokens. \labrador optimized a simple
multi-task loss that was the sum of its categorical cross-entropy for
lab code prediction and mean-squared error (MSE) for lab value
prediction. 

%% file: sections/results.tex
In the following sections, we assess \labrador and BERT on intrinsic evaluations to understand the extent to which each model accomplishes the pre-training task. Following that, we fine-tune both models for downstream outcome prediction on four datasets: COVID-19 diagnosis, cancer diagnosis, and the prediction of sepsis-related mortality and alcohol consumption level. 

\subsection{Intrinsic evaluations}\label{intrinsic-evaluations}

\subsubsection{Assessment of pre-training loss
}\label{assessment-of-pre-training-loss}

BERT and \labrador were pre-trained until each converged in
their pre-training loss (Figure \ref{fig_labrador_loss} and \ref{fig_bert_loss}). After $1.5 \times 10^6$ steps, BERT reaches
a validation perplexity of 1.3 tokens, whereas after $5 \times 10^5$ steps \labrador reaches a validation perplexity of 1.02 tokens and MSE of 0.013,
respectively. Recall that BERT and \labrador have vocabulary sizes of 4251 and 531 tokens, respectively. This shows that both models clearly master their categorical prediction tasks.

Also, \labrador's final validation MSE of 0.013 suggests that its average error is $\pm \sqrt{0.013} \approx \pm 0.114$, which corresponds approximately to quintile resolution in the lab value distribution because \labrador's lab values are uniform on the interval {[}0, 1{]} (described in Section \ref{preprocessing-of-the-pre-training-lab-data}).

\subsubsection{Embedding space
visualization}\label{embedding-space-visualization}

Despite the fact that BERT and \labrador have comparable pre-training perplexity,
the structure of their embedding space differs greatly. We
performed dimensionality reduction on BERT and \labrador's embeddings for the labs in the test split using the UMAP
algorithm \citep{mcinnes2018umap}. Section \ref{umap} describes the details of the experiment. Figure \ref{fig_umap} presents the two-dimensional structure of the 70 most frequent lab tests, which are colored according to the
panel of labs that they are most often ordered with. Appendix \ref{appendix_labpanels} defines each lab panel.

\labrador has a well-separated cluster of embeddings for each lab code,
whereas BERT has far less separation. Qualitatively, we do not observe any
clinical meaning to the relative distances between each pair of
\labrador's embedding clusters. In contrast, the relative positioning of the large embedding islands in BERT's UMAP plot may possess some basic clinical meaning. For instance, urinalysis tests (pink) are directly
adjacent to urine toxicology screening tests (light blue). Similarly,
the CBC tests (dark blue) form a continuation with the CBC with
differential tests (purple). We note that the two most common panels of
lab tests (CBC and BMP) do not appear to be well-separated in BERT's embedding space, whereas they are completely separated in
\labrador's embedding space. 

Panel B of Figure \ref{fig_umap} visualizes the embedding space for the
four most frequently ordered lab tests and colors these embeddings by
their lab value. While \labrador learns a smooth
gradient representing the quantitative value for each lab test,
BERT does not. There is some directionality in the BERT embedding
space that correlates with lab value, but it is notably less monotonic
when compared to \labrador. 

It is important to note that while low-dimensional
representations of embeddings can be helpful in providing high-level, qualitative intuitions for how a model behaves, they can be misleading and unfaithful to the original, high-dimensional
space \citep{chari2023specious}.
We provide these visualizations to provide some insight into the
different ways \labrador and BERT learn to represent the input space, but
caution should be used when attempting to make fine-grained,
quantitative conclusions.

\subsubsection{Imputation}\label{imputation}

We also evaluated \labrador and BERT for their ability to impute missing
lab values in the test split of their pre-training data. To do so, we mask a randomly selected lab value
from each bag of labs in the test split and run each model's forward pass to obtain its prediction for the masked value (see Section \ref{intrinsic-imputations} for more details). In panel A of Figure \ref{fig:imputations}, we see that the predictions from both pre-trained models achieve a Pearson correlation $r^2 > 0.8$. As an ablation,
we also evaluated \labrador and BERT with randomly initialized
parameters to assess the effect of pre-training on the imputations. We see that the strong correlation between the imputed and true lab values
stems entirely from pre-training and is not an artifact of the model architectures or the data.

In panel B of Figure \ref{fig:imputations}, we present individual scatter plots for the four best lab
tests as measured by Pearson correlation. For \labrador, there is a
clear relationship between the strength of the correlation and
the frequency of the lab test in the training data. Interestingly, we found that this
relationship was less pronounced for BERT, as there were some frequent
lab tests where the model performed relatively poorly and a small number
of rare tests where the model performed well. In panel C, we show the
four worst lab tests as measured by Pearson correlation. It should be
noted that although the majority of MIMIC-IV laboratory data consists of
common test results, such as the tests from the panels shown in Figure \ref{fig_umap}, there is a
long tail of rare tests in the database. \labrador and BERT may be less capable of capturing the statistical
dependencies in this long tail.

\subsection{Extrinsic evaluations}\label{extrinsic-evaluations}

In this section, we evaluate \labrador and BERT for their ability to improve predictive performance in real-world, downstream use cases. We also conduct an ablation study to understand which parts of each model contribute most to its performance.

\subsubsection{Fine-tuning experiments and outcome
prediction}\label{fine-tuning-experiments-and-outcome-prediction}

We compare \labrador and BERT's performance to a set of baseline models in four
distinct outcome scenarios where predictions were made on the basis of
laboratory measurements: ICU mortality due to sepsis, cancer diagnosis,
COVID-19 diagnosis, and the prediction of alcohol consumption. Baseline methods are defined in Section \ref{baseline-methods-for-outcome-prediction} and pre-processing for each evaluation dataset is described in Section \ref{evaluation-data-preprocessing}.

For each of these experiments, we add a prediction head to \labrador or BERT consisting of multiple fully-connected layers followed by
extensive hyperparameter tuning for this prediction head on each outcome
dataset and with each base model. Section \ref{transformer-fine-tuning} describes this fine-tuning procedure in detail. Table \ref{table_outcomeprediction} shows the performance of
\labrador and BERT compared to simple logistic (or linear) regression,
Random Forest and XGBoost. 

XGBoost narrowly outperforms all other
methods on sepsis mortality prediction, cancer diagnosis and COVID-19
diagnosis. \labrador performs best on alcohol consumption prediction,
which is a regression problem (as opposed to the other 3 classification
tasks) and is also the smallest outcome dataset. Additionally, \labrador
outperforms BERT across all four evaluations, suggesting there is value
in a model that explicitly accounts for the continuous nature of the
data. For a precise definition of each evaluation task see Section \ref{evaluation-data-preprocessing}. 

\begin{table*}[h]
\centering
\begin{threeparttable}
\caption{Comparison of predictive performance among \labrador, BERT, and baseline models across four outcome prediction settings.}
\label{table_outcomeprediction}
\small
\begin{tabular}{l|cccc}
\toprule
& \parbox[t]{2.6cm}{\centering Sepsis \\ mortality ($\downarrow$)} & \parbox[t]{2.6cm}{\centering Cancer \\ diagnosis ($\downarrow$)} & \parbox[t]{2.6cm}{\centering COVID-19 \\ diagnosis ($\downarrow$)} & \parbox[t]{2.6cm}{\centering Alcohol \\ consumption ($\downarrow$)} \\
\midrule
LR\tnote{1} & 0.410 (0.398, 0.424) & 1.177 (1.100, 1.272) & 0.462 (0.434, 0.529) & 9.73 (4.83, 13.93) \\
RF & \textbf{0.387} (0.372, 0.401) & 0.942 (0.902, 0.987) & 0.431 (0.404, 0.487) & 9.65 (4.62, 13.35) \\
XGB & \textbf{0.387} (0.376, 0.399) & \textbf{0.918} (0.847, 0.989) & \textbf{0.419} (0.371, 0.504) & 10.84 (5.54, 15.39) \\
\midrule
BERT & 0.406 (0.368, 0.431) & 1.01 (0.859, 1.127) & 0.441 (0.391, 0.513) & 8.34 (5.80, 12.71) \\
\labrador & 0.400 (0.371, 0.420) & 0.978 (0.889, 1.163) & 0.425 (0.369, 0.487) & \textbf{7.11} (5.33, 9.15) \\
\bottomrule
\end{tabular}
\begin{tablenotes}
\small
\item Note: We report the metrics as mean (min, max) across 5 random replicates. The best model for each evaluation dataset is in bold. We use cross-entropy as evaluation metrics for all but alcohol consumption, where we use MSE. Lower values indicate better predictive performance in all cases.
\item[1] LR = logistic regression for all evaluations but alcohol consumption, where LR = linear regression.
\end{tablenotes}
\end{threeparttable}
\end{table*}

\subsubsection{Ablation study}\label{ablation-study}

It is important to differentiate the inductive bias of the model
architecture from the effect of pre-training, since previous studies
have shown that many architectures (including Transformers) can perform surprisingly well
even before they are trained on any data
\citep{Rives2021-ou}. In
order to assess this, we perform an ablation study where the
parameters of BERT and \labrador are randomly initialized and undergo the same hyperparameter tuning as their pre-trained counterparts. The difference between a pre-trained model's performance and its
ablation performance isolates the effect of pre-training on these
downstream outcome prediction tasks. In Table \ref{table_ablation}, we observe that the pre-trained \labrador outperforms its ablation in 3 of 4 downstream tasks with the exception of cancer diagnosis. In contrast,
the pre-trained BERT underperforms its ablation in 3 of 4 downstream tasks with the exception of COVID-19 diagnosis. Table \ref{table_ablation} demonstrates that \labrador's transfer learning is strictly superior to BERT's.

\begin{table*}[ht]
\centering
\begin{threeparttable}
\caption{Ablation study evaluating the influence of pre-training on the predictive performance of BERT and \labrador in four outcome prediction settings.}
\label{table_ablation}
\small
\setlength{\tabcolsep}{1.25pt}
\begin{tabular*}{\textwidth}{@{\extracolsep\fill}lcc|cc}
\toprule
\multicolumn{1}{l}{}
& \multicolumn{2}{c}{BERT} & \multicolumn{2}{c}{\labrador} \\
\cmidrule(lr){2-3} \cmidrule(lr{10pt}){4-5}
\multicolumn{1}{l}{}
 & \raisebox{-1ex}{Pre-trained} & \multicolumn{1}{c}{\parbox[t]{2.5cm}{\centering Random \\ weights (\%)\footnotemark[1]}} & \raisebox{-1ex}{Pre-trained} & \parbox[t]{2.5cm}{\centering Random \\ weights (\%)\footnotemark[1]} \\
\midrule
Sepsis mortality & 0.406 (0.368, 0.431) & \textbf{+0.5\%} & \textbf{0.400} (0.371, 0.420) & -2\% \\
Cancer diagnosis & 1.01 (0.859, 1.127) & \textbf{+8.8\%} & 0.978 (0.889, 1.163) & \textbf{+5.1\%} \\
COVID-19 diagnosis & \textbf{0.441} (0.391, 0.513) & -3.2\% & \textbf{0.425} (0.369, 0.487) & -3.8\% \\
Alcohol consumption & 8.34 (5.80, 12.71) & \textbf{+11.5\%} & \textbf{7.11} (5.33, 9.15) & -5.9\% \\
\bottomrule
\end{tabular*}
\begin{tablenotes}
\small
\item[1] A positive percentage corresponds to an improvement in the evaluation metric relative to the corresponding pre-trained model.
\end{tablenotes}
\end{threeparttable}
\end{table*}

%% file: sections/discussion.tex
    Through extensive experiments on downstream tasks, we demonstrated that \labrador consistently outperforms the standard BERT architecture on both intrinsic and extrinsic evaluations. Our results show the value of an architecture specialized for laboratory data compared to the traditional approach of discretizing numeric values for tokenization. A key contribution of our work is the creation of an effective self-supervised objective for pre-training on the challenging distribution of laboratory data in EHRs. 

    Both \labrador and BERT were successfully optimized as masked language models, learning useful representations of laboratory tests. This was evidenced by their ability to accurately impute missing values in the test set. However, our fine-tuning experiments reveal that transfer learning from pre-training on laboratory data has limited
    success. On outcome prediction tasks, neither \labrador nor BERT consistently surpass simple baseline methods. Our ablation study (Section \ref{ablation-study}) indicates that this stems from insufficient representation learning
    during pre-training, rather than issues with the inductive biases or fine-tuning.

    Next, we posit several reasons why pre-training on laboratory data may fail to produce significant downstream gains, and provide a summary of the available support for each.
    
    \noindent\textbf{Lab data lack correlation structure}. If lab values were statistically independent then the pre-training task would be impossible. Likewise, if there is a trivial correlation structure (e.g. some pairs of labs are perfectly correlated) then we would not expect an MLM objective to outperform simple models. However, we believe that there is a meaningful and complex correlation structure amongst the lab measurements, as evidenced in Figure \ref{fig_corrmatrix}.
    
    \noindent\textbf{MLM masking rate is too low.} \labrador and BERT may satisfy the pre-training objective without learning enough about labs and patients to be adept at outcome prediction. During our pre-training, the average mask rate is just 5\% per bag. Unfortunately, increasing the mask rate may be impossible because each bag of labs is permutation invariant. As a result, if multiple elements are masked within a bag, \labrador and BERT's predictions for each mask token will be identical. There is no way for the model to distinguish one mask token from the
    next in a permutation invariant setting.
    
    Positional embeddings are typically used to break this symmetry, but we are not aware of a variable to encode using via positional embeddings. For example, the index in the input cannot be used since the relative ordering of multiple mask tokens is arbitrary. The timestamp when a lab
    was measured could be used but labs are ordered in sets at the exact same moment in time in a patient's medical record. Even if a bag of labs includes tests that span multiple order sets, there would still be a symmetry issue
    within each order set. We believe that this is a general challenge for any MLM of electronic health record data, since measurements are often made in order sets.
    
    \noindent \textbf{Pre-training and downstream evaluations are a poor match.} The set of features used for cancer diagnosis is the least similar to the pre-training vocabulary and is also the evaluation that both
    \labrador and BERT perform the worst on. However, sepsis mortality prediction uses lab data from the same ICU data distribution as in pre-training and yet it is not the evaluation where pre-training delivers the most value, which would be COVID-19 diagnosis. Therefore, we do not believe that a mismatch between pre-training and downstream
    data distributions is the primary factor explaining the failure in transfer learning.
    
    \noindent\textbf{Lab data only partially capture patient state}. Lab data, unlike text, do not provide a comprehensive view of a patient\textquotesingle s condition, necessitating the incorporation of additional patient data categories. Thus, it could be that even simple
    models are close to the Bayes error rate given the information captured by lab measurements. We propose that future work should jointly learn from multiple EHR data categories, like diagnostic codes, procedure codes, and medication prescriptions. This could lead to a multimodal model predicting next tokens across structured EHR data, medical images, and clinical text.
    
    \noindent\textbf{Insufficient data scale.} The capabilities of frontier LLM's have emerged as a function of
    the scale of the pre-training data. We pre-trained \labrador and BERT on approximately 100 million lab tests but this only corresponds to 4.5 million input samples. In comparison,
    GPT-3 \citep{brown2020language} was trained on a corpus of
    300 billion tokens where each input sequence contained 2048 tokens or about 145 million input sequences. This is nearly two orders of magnitude larger than our pre-training dataset in terms of input sequences and over 3 orders of magnitude larger in token count. We believe that this is at least 1
    order of magnitude larger than the largest available lab dataset today. Therefore, we encourage others to contribute to the harmonization of datasets across many sources, which requires large organized coding efforts to unify the data into a compatible vocabulary. This also requires open access to such data sources, as MIMIC has provided. Importantly, if we take into account the Chinchilla scaling laws from \cite{hoffmann2022empirical}, it is plausible that scaling model size will not yield significant benefits in downstream performance given the relative scarcity of lab data compared to text.

%% file: sections/conclusion.tex
In this work, we presented \labrador, a novel Transformer architecture that can be pre-trained on (integer, float) data such that the
float values are treated in a continuous manner. We showed that
\labrador outperforms BERT across all downstream evaluations and appears
to learn monotonic axes corresponding to lab value in its embedding
space. However, despite the success seen in pre-training, both models
fail to outperform XGBoost on four outcome prediction tasks. Our
ablation studies reveal that transfer learning largely fails in this
setting despite being more successful for \labrador than for
BERT. We hypothesize that this failure is largely due to a lack of
pre-training data scale and an insufficient characterization of the data generating process underlying each patient using
labs in isolation from other clinical data types. As a result, we
urge future work to focus on harmonizing disparate datasets and jointly modeling several categories of EHR data.

A persistent challenge in medical machine learning for structured data is the lack of universally accepted evaluations. Model predictions are difficult to verify, since there is no human baseline for performance. As a result, we have a poor sense of the Bayes error rate for these types of prediction tasks. We argue that a prerequisite to progressing the self-supervised learning efforts on EHR data is the establishment of universally accepted evaluations that will provide a clear picture of the field's progress over time in comparison to tree-based methods \citep{Bellamy2020-qr}. 

%% file: sections/appendix.tex
\section{Background: transformers and masked language models}\label{appendix_background}

\subsection{Transformers}

Transformers are a class of deep learning models introduced by \cite{Vaswani2017-hn} that have significantly impacted various areas of machine learning, including natural language processing, computer vision, and reinforcement learning. Transformers are based on the principle of self-attention, which allows them to efficiently capture long-range dependencies in input data by computing weighted combinations of all input elements instead of relying on the fixed-size sliding windows used in traditional convolutional and recurrent neural networks.

The core building block of a Transformer is the attention mechanism, which computes the similarity between all pairs of input elements and replaces each input element with a weighted average of all other inputs, weighted by their similarity. Transformers are built by stacking multiple layers of self-attention mechanisms, enabling the model to learn complex patterns and representations in the input data.

\subsection{Masked language models}

Masked language models (MLMs) are a popular pre-training technique for
natural language processing tasks. The key idea behind MLMs is to train
a model to predict missing words in a given text, with some portion of
the input words being masked out. By learning to predict the masked
words, the model is forced to capture the underlying structure and
semantics of the language, resulting in a more robust and generalizable
representation.

MLMs are usually pre-trained on large-scale text corpora and fine-tuned
on specific downstream tasks, such as text classification, sentiment
analysis, or named entity recognition. This two-step process, consisting
of pre-training and fine-tuning, is known as transfer learning.
Pre-training on large corpora allows the model to learn general language
features, while fine-tuning adapts the model to the nuances of the
specific task at hand.

The success of MLMs is primarily attributed to their ability to learn
contextualized word representations, which capture both syntactic and
semantic information from the input text. This is in contrast to
traditional word embedding techniques, such as Word2Vec \citep{mikolov2013efficient} and GloVe \citep{Pennington2014-ut}, which
learn static word representations that are context-independent.

In this work, we leverage the power of Transformers and the masked
language model pre-training technique to explore their applicability to
laboratory data in medicine. We aim to assess whether the successes of
Transformers and MLMs in other domains can be replicated in the context
of medical data, particularly in outcome prediction tasks based on
laboratory data.

\section{Pseudo-code Implementations}
\label{pseudocode}

The continuous embedding layer is implemented as follows: 

\begin{verbatim}
def ContinuousEmbeddingLayer(
    lab_values, token_embeddings):
    x = Dense_pw(lab_values, 
        out_size=d_model, 
        activation=`linear')
    x = x + token_embeddings
    x = Dense_pw(x, out_size=d_model, 
        activation=`relu')
    x = LayerNorm(x)
    return x
\end{verbatim}

Where \verb|Dense_pw| represents a position-wise dense layer and \verb|token_embeddings| is the output of the categorical embedding layer.

The continuous prediction head is implemented as follows:

\begin{verbatim}
def ContinuousPredictionHead(
    final_layer_embeddings, probs):
    x = concat([final_layer_embeddings, 
                probs])
    x = Dense_pw(x, 
        out_size=embed_dim + probs_dim, 
        activation=`relu')
    x = Dense_pw(x, out_size=1, 
        activation=`sigmoid')
    return x
\end{verbatim}

Where \verb|probs| is the categorical head's probability distribution over the lab code vocabulary. This supplies the continuous prediction head with each token's predicted probability when attempting to impute the masked lab value.

\newpage
\onecolumn
\section{Embedding Space Visualization}
\begin{figure*}[htbp]
\centering
\includegraphics[width=\textwidth]{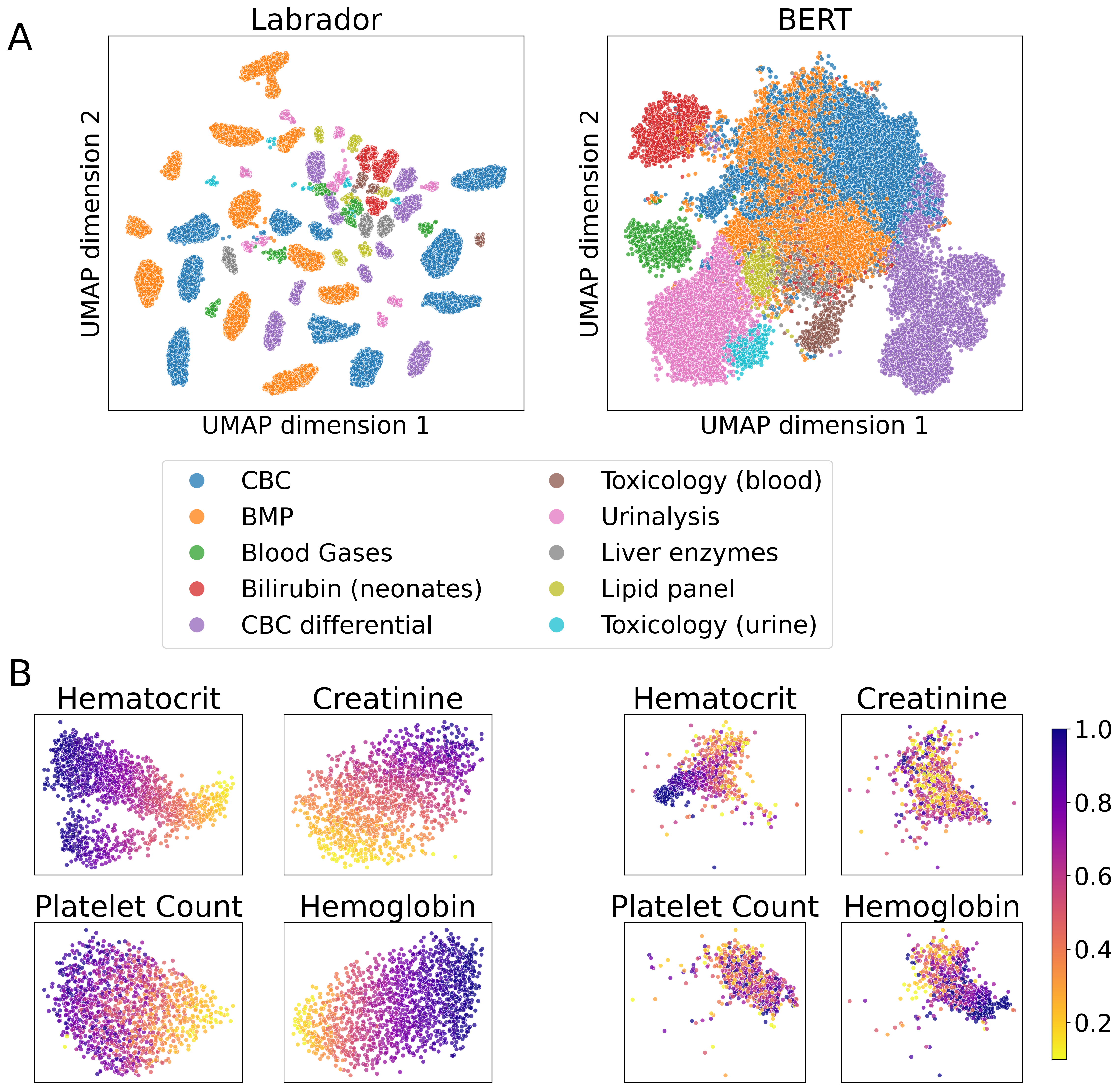}
\caption{UMAP Visualization of \labrador and BERT Embeddings.
\textbf{A.} A global view of the embedding space structure for \labrador (left)
and BERT (right). The 70 most frequently ordered lab tests are
shown and colored according to the panel of tests they are typically
ordered with (see Appendix \ref{appendix_labpanels} for all panel definitions).
All labs on this figure are from the test split and were not seen during
pre-training. \textbf{B.} Visualization of embeddings for four routinely
collected laboratory measurements colored by lab value and scaled to the interval {[}0, 1{]}. \labrador appears to encode the measured lab value in a
much more natural way with a smooth gradient for lab value compared to BERT.}
\label{fig_umap}
\end{figure*}

\newpage
\section{Imputation Evaluation of Labrador and BERT}
\begin{figure*}[htbp]
\centering
\includegraphics[width=.99\linewidth]{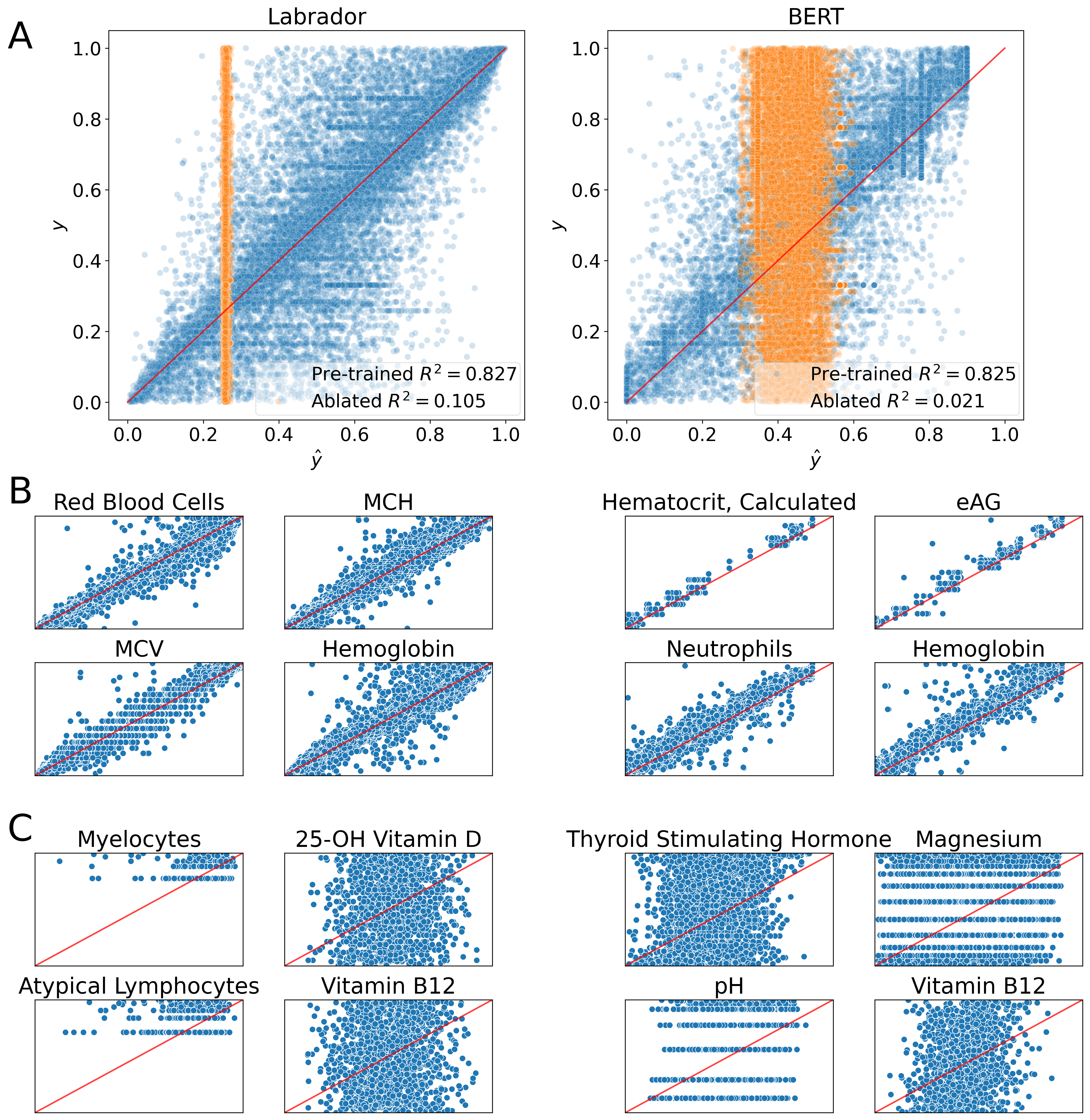}
\caption{Intrinsic Evaluation of \labrador and BERT by lab value
imputation. \textbf{A.} We masked a random lab from each bag in the test set of the
pre-training data and imputed these values using
pre-trained \labrador (left) and BERT (right). Both \labrador and BERT
achieve a Pearson correlation $r^2 > 0.8$, in contrast to their
ablations (orange). \textbf{B.} Imputations for the four best lab tests as
measured by Pearson correlation. \textbf{C.} Imputations for the four worst lab tests as measured by Pearson correlation.}
\label{fig:imputations}
\end{figure*}

\newpage
\section{Parameter-matched BERT experiments}\label{appendix_parameter_matched}
\begin{figure*}[h!]
\centering
\includegraphics[width=.75\linewidth]{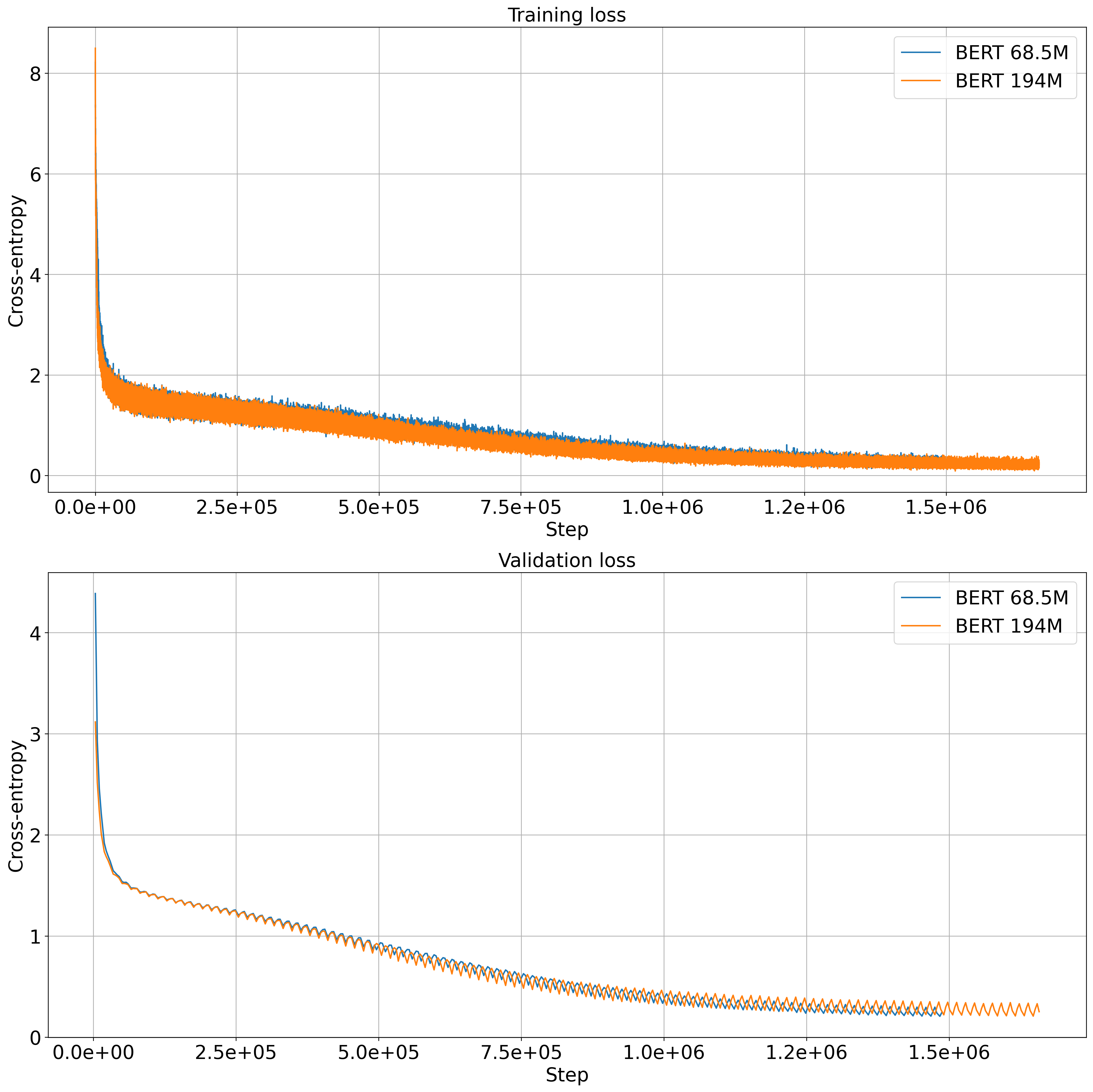}
\caption{Pre-training loss for BERT with 68.5M ($d_k = \left\lfloor \frac{d_{\text{model}}}{h}\right\rfloor$) versus 194M parameters ($d_k = d_{\text{model}}$). \textbf{Top:} The training loss for both BERT models. \textbf{Bottom:} The validation loss.}
\label{fig_bert_loss_comparison}
\end{figure*}
To ensure that our choice for \labrador's key dimension, $d_k = d_{\text{model}}$ (instead of the more common $d_k = \frac{d_{\text{model}}}{h}$, where $h$ is the number of attention heads) does not alter our results or conclusions, we pre-trained a second BERT model with the same $d_k$ as \labrador and repeated our evaluations. 

\begin{figure*}[htbp]
\centering
\includegraphics[width=.75\linewidth]{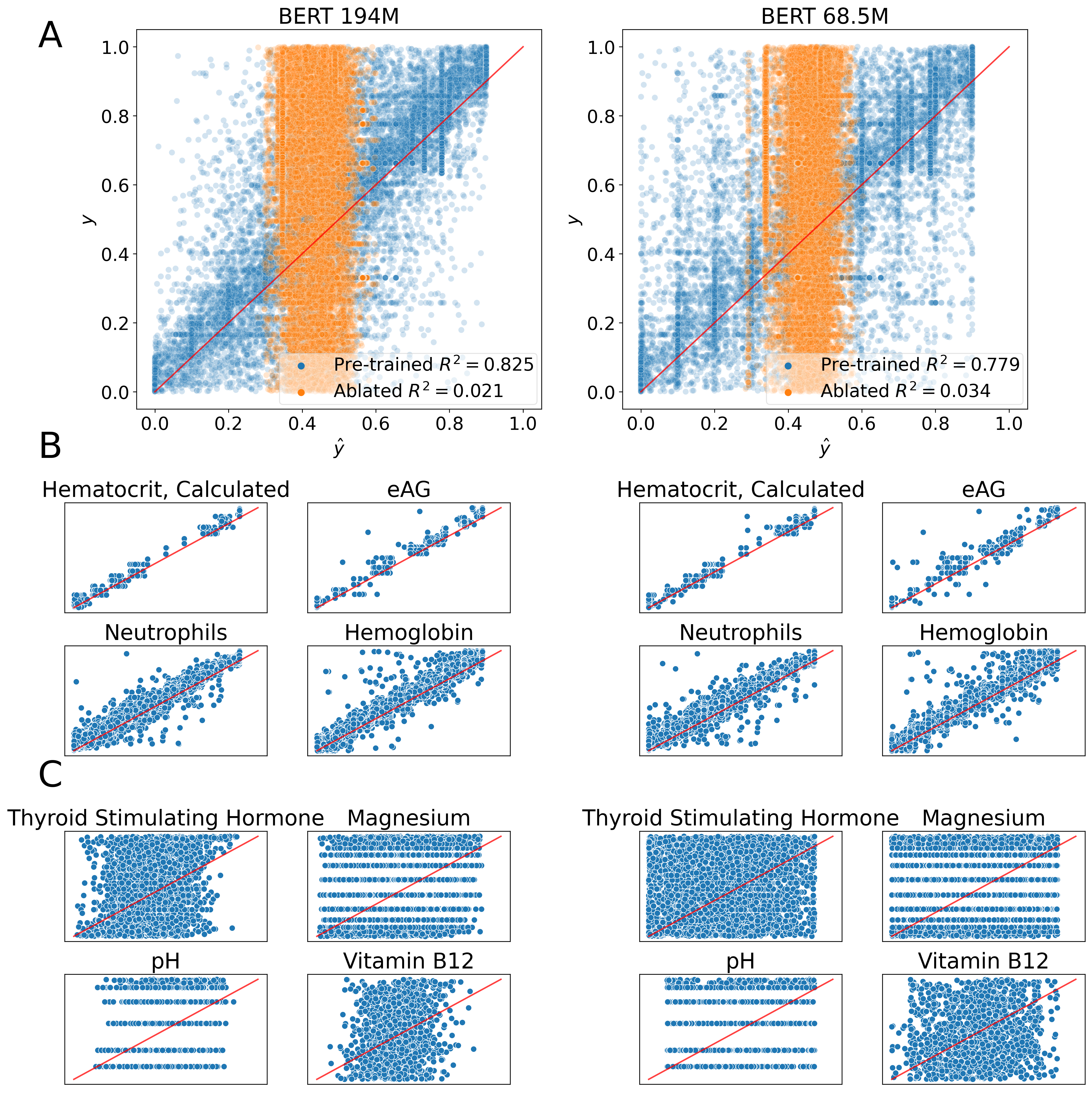}
\caption{Lab value imputations from BERT with 68.5M ($d_k = \left\lfloor \frac{d_{\text{model}}}{h}\right\rfloor$) versus 194M parameters ($d_k = d_{\text{model}}$). \textbf{A.} Imputations from both pre-trained BERT models (blue) as well as their ablations (orange) on the test split of the pre-training data. \textbf{B.} Imputations for the four best lab tests as measured by Pearson correlation. \textbf{C.} Imputations for the four worst lab tests as measured by Pearson correlation.}
\label{fig_bert_imputation_comparison}
\end{figure*}

BERT with $d_k = \frac{d_{\text{model}}}{h}$ has 68.5M parameters, whereas the BERT model with the larger $d_k$ has 194M parameters. The latter matches \labrador's parameter count of 196M as closely as possible, since their backbone structure is identical besides the additional 2M weights in \labrador's continuous embedding layer and continuous prediction head. Figure \ref{fig_bert_loss_comparison} compares the pre-training loss curves for BERT with 68.5M versus 194M parameters. Figure \ref{fig_bert_imputation_comparison} compares their intrinsic lab value imputation performance. The imputation performance of BERT 194M is slightly better than BERT 68.5M, despite their pre-training loss curves appearing nearly identical. 

Most importantly, we also include the larger BERT's fine-tuning performance on the downstream outcome prediction tasks, both with pre-training and without. Table \ref{table_compare_bert_extrinsic} shows that the smaller BERT model outperforms the larger BERT on 7 out of the 8 evaluations, but in general the difference is not statistically significant.


\begin{table*}[htbp]
\caption{Comparison of fine-tuning performance between BERT with 68.5M versus 194M parameters. \textbf{Left:} Performance of each model size after pre-training. \textbf{Right:} Performance of each model size starting from randomly initialized weights.}\label{table_compare_bert_extrinsic}
\small
\setlength{\tabcolsep}{1.25pt}
\begin{tabular*}{\textwidth}{@{\extracolsep\fill}lcc|cc}
\toprule
\multicolumn{1}{l}{}
& \multicolumn{2}{c}{Pre-trained} & \multicolumn{2}{c}{Random weights} \\
\cmidrule(lr){2-3} \cmidrule(lr{10pt}){4-5}
 & BERT 68M ($\downarrow$) & BERT 194M ($\downarrow$) & BERT 68M ($\downarrow$) & BERT 194M ($\downarrow$) \\
\midrule
Sepsis mortality & 0.406 (0.368, 0.431) & \textbf{0.400} (0.368, 0.424) & \textbf{0.404} (0.378, 0.418) & 0.406 (0.379, 0.419) \\
Cancer diagnosis & \textbf{1.01} (0.859, 1.127) & 1.08 (0.978, 1.20) & \textbf{0.921} (0.774, 1.054) & 0.972 (0.884, 1.10) \\
COVID-19 diagnosis & \textbf{0.441} (0.391, 0.513) & 0.484 (0.400, 0.544) & \textbf{0.455} (0.412, 0.511) & 0.470 (0.435, 0.489) \\
Alcohol consumption & \textbf{8.34} (5.80, 12.71) & 8.68 (6.57, 12.41) & \textbf{7.38} (4.93, 10.43) & 7.49 (4.39, 11.28) \\
\bottomrule
\end{tabular*}
\end{table*}

\twocolumn
\section{Further experimental details}\label{appendix_further_exp_details}
\subsection{UMAP}\label{umap}

After performing several scaling experiments with UMAP reduction on the
full test split of the pre-training data, we observed that the results
were not significantly altered if we used a random sub-sample of the
data. We randomly selected bags of labs totalling 50,000 (lab code, lab
value) pairs from the test split. These bags separately underwent
\labrador and BERT's forward pass to obtain the final layer embeddings
from each Transformer. The UMAP algorithm was run with \verb|n_neighbors=600|,
\verb|min_dist=0.99| and a random seed of 3141592. In preliminary experiments,
we found that increasing the number of neighbors and increasing the
minimum distance between points in the reduction provided a better
global structure of the embedding space. In consultation with the MIMIC
database authors, we manually labeled lab codes according to their panel
membership, such as the Complete Blood Count (see Appendix \ref{appendix_labpanels} for all
panels). This allowed us to color Figure \ref{fig_umap} accordingly.

\subsection{Intrinsic imputations}\label{intrinsic-imputations}

For intrinsic imputations, we randomly sampled approximately 2 million
bags of labs from the pre-training test split and randomly masked one
(lab code, lab value) pair. These masked bags were run through the
forward pass of \labrador and BERT in order to assess their predictions
for the masked lab value. \labrador's continuous prediction head outputs a predicted lab value, which we compared directly to the true masked lab value.

Since BERT only
predicts tokens, we experimented with two methods for converting these
categorical predictions into continuous values. First, we tried an
argmax method, which simply extracts the token with the greatest logit
in BERT's output distribution and set the predicted lab value equal to the lower
bound of that token's defined decile. Second, we tried a weighted
quantile method, which takes a weighted average of the lower bounds of
each token's decile where the weights are the re-normalized probabilities for that lab code's subset of the output distribution. We found that this
weighted quantile method works far better than the argmax method, which
is likely due to the fact that the model is able to hedge its
probabilities across several deciles.

We present the Pearson correlation between the true lab
values and each model's predictions in Figure \ref{fig:imputations}. We also calculated Pearson
correlations for each unique lab code and used this to rank lab codes
from best to worst in terms of their imputation accuracy. Finally, we performed an ablation experiment to highlight what was learned during
pre-training. The ablation was conducted by randomly initializing
the weights for each model prior to performing the steps described
above.

\subsection{Pre-processing the evaluation data}\label{evaluation-data-preprocessing}

\subsubsection{COVID-19 cohort}\label{covid-19-cohort}

For COVID-19 diagnosis, we used a public dataset \citep{cabitza2020dataset, cabitza2021development} that contains blood
work for 1,624 patients (52\% COVID-19 positive) admitted at San Raphael
Hospital from February to May 2020. The data contain 34 features and a
binary target variable for COVID-19 diagnosis determined by RT-PCR at a
later point in time. Features `CK' (creatine kinase) and `UREA' (blood
urea nitrogen) were dropped from the data because their rates of
missingness were too high (60\% and 38.5\%, respectively). Any patients
that were missing greater than 25\% of the remaining 32 features were
dropped as well since we believe that the imputation quality for these
patients would be too poor. This resulted in 1,312 patients.

The features contain 26 lab tests that exist in the MIMIC-IV vocabulary
of \labrador and BERT. We manually verified the scale of each lab test to
ensure that the units of measurement were identical. One test, `CA'
(calcium), was initially in mmol/L so we converted it to mg/dL, which is
the unit of measurement used in MIMIC-IV for this test (itemid 50893). Our pre-processed version of the COVID-19 dataset can be found in our codebase on GitHub.

During evaluation of the baseline methods, any missing values were
imputed using the column mean in the current split during k-fold
cross-validation. In contrast, both Transformers have natural ways of
handling missing values without requiring imputation. \labrador uses a
special null token for missing values and BERT has a special token in
its vocabulary for missing values for each lab code. Finally, we used
the stored eCDFs from pre-training to convert all lab values from the 26
in-vocabulary features to the interval $[0,1]$. \labrador was
fine-tuned directly on these values. For BERT, an additional step is
required which maps the eCDF values to their respective tokens in the
model's vocabulary.

\subsubsection{Sepsis cohort}\label{sepsis-cohort}

We downloaded the MIMIC-IV sepsis cohort from Google BigQuery under \path{physionet-data/mimic_derived/sepsis3}. Accessing the MIMIC-IV data \citep{johnson2020mimic, johnson2023mimic} requires signing a data use agreement and completing a short online training, so we cannot share this dataset directly. However, after obtaining MIMIC-IV access on PhysioNet, you can find instructions on accessing the data using Google BigQuery. 

The sepsis dataset contains information on
34,678 septic patients, such as their patient ID, hospital stay ID, and
several other variables. The variable \verb|sofa_time|, which is the timestamp for the patient's most recent SOFA score, was used
to link patients to their other MIMIC records. These patients were
linked to their admission record to determine whether they survived their ICU stay or not. We created a binary outcome variable equal
to 1 if the patient died by the end of their stay and 0 otherwise. We
also linked patients to their lab test results. We filtered the labs to
only include those that were measured in the first 24 hours of the
patient's stay in the ICU. Patients with fewer than 2 lab tests during the
first 24 hours were dropped from the analysis, which left 34,136 remaining patients.

In order to conduct a fair
comparison with baseline methods, we opted to create a tabular
dataset out of this ragged dataset of lab tests. To do so, we selected
the 23 most frequently occurring lab tests in the first 24 hours. We
chose the top 23 because this produced a tabular dataset for all 34,136
patients with under 5\% missingness. Finally, we converted all lab
values to the interval $[0,1]$ using the eCDF method described
in Section \ref{preprocessing-of-the-pre-training-lab-data} and mapped these values into BERT's token
vocabulary. The code to recreate this dataset is included in our codebase.

\subsubsection{Cancer diagnosis cohort}\label{cancer-diagnosis-cohort}

For cancer diagnosis, we used a public dataset shared by \cite{tsai2022machine} that contains blood work for 1,336 patients. Briefly, each patient
received a diagnosis of either cystitis, bladder cancer, kidney cancer,
uterine cancer, or prostate cancer. The original dataset contained 39
features in addition to the 5-class target variable. Five of these 39
features were removed for having too much missingness (urine epithelium,
A/G ratio, urine ketone, urine glucose, and strip WBC). Among the
remaining 34 features, 24 were lab tests within the MIMIC-IV vocabulary,
2 lab tests were out of vocabulary, and there were 8 social variables
(age, gender, smoking status, diabetes status, etc.). We manually
verified that each in-vocabulary lab test was using the same units of
measurement as in MIMIC-IV. For categorical variables that had missing
values, we used the missing indicator method. We transformed all lab
values using the eCDF method and missing lab values were handled in the same manner as for the COVID-19 dataset (Section \ref{covid-19-cohort}).

\subsubsection{Alcohol consumption
cohort}\label{alcohol-consumption-cohort}

We obtained the alcohol consumption dataset from the UCI Machine Learning
Repository\footnote{\url{https://doi.org/10.24432/C54G67}}. This small dataset contains 345 patients, 5 lab test features and 1
continuous outcome variable. Briefly, the outcome is the number of
half-pint equivalents of alcoholic beverages drunk per day. The 5 lab
test features are all thought to be sensitive to liver disorders that
might arise from excessive alcohol consumption. There is no missingness
in this dataset therefore no pre-processing was required for the baseline
methods. For \labrador and BERT, the lab values were transformed with the
eCDF method.

\subsection{Transformer fine-tuning}\label{transformer-fine-tuning}

In order to fine-tune \labrador and BERT on the four downstream outcome
prediction tasks, we created a simple wrapper class for each model that
could handle a combination of in-vocabulary lab features as well as
previously unseen features. The pseudocode for \labrador's wrapper class'
call method is as follows:

\begin{verbatim}
def LabradorFinetuneWrapper(
    inputs: Dict[Tensor]) -> Tensor:
    x = base_model(inputs['lab_features'])
    x = pool(x)
    
    if 'non_mimic_features' in inputs.keys():
        non_mimic_features = dense(
            inputs['non_mimic_features'])
        x = concat([x, non_mimic_features], 
                   axis=1)
    
    x = dense(x)
    x = dropout(x)
    x = dense(x)
    return x
\end{verbatim}

The lab features are passed to the base model as usual. The base model's
final embeddings are averaged into a single vector of size
\verb|hidden_dim| (i.e. 1024 in our experiments). If there are other
features, these are passed through a 1-to-1 dense layer with a ReLU
activation followed by concatenation with the pooled embedding vector.
This combined representation is passed through an additional 1-to-1
ReLU-activated dense layer prior to the final dense layer that projects
this vector down to the size required by the outcome prediction task.
This final dense layer also has an activation function that is dependent
on the outcome prediction task (e.g. linear for alcohol consumption
prediction, sigmoid for sepsis mortality prediction). The
pseudocode is the same for BERT.

Across all fine-tuning experiments, we froze the parameters of the base
model and performed hyperparameter tuning across a grid consisting of the
learning rate, dropout rate, number of fine-tuning epochs and batch
size. More specifically, we use the following grid of hyperparameters:

\begin{itemize}
    \item Number of epochs: $[30, 60, 90]$
    \item Batch size: $[16, 32, 64]$
    \item Learning rate: [1e-4, 3e-4, 5e-4, 1e-3]
    \item Dropout rate: $[0.1, 0.3, 0.5, 0.7]$
\end{itemize}

In our preliminary experiments, we found that this grid spans all
plausible regions of interest in hyperparameter space. We also conducted
preliminary experiments with unfrozen base models and the same
hyperparameter grid as the original authors of BERT, but we found that this
resulted in worse performance.

\subsection{Baseline methods for outcome
prediction}\label{baseline-methods-for-outcome-prediction}

We include a set of simple baseline
methods for comparison in our fine-tuning experiments. This included linear regression, logistic
regression, Random Forest (classification and regression) and XGBoost
(classification and regression). For
all logistic regression experiments, we tuned $C$, the coefficient for the
strength of $L_2$-regularization, using the grid: {[}0.0001, 0.001, 0.01,
0.1{]}. Otherwise, we allowed 200 iterations for the solver to converge
and used the lbfgs solver. For linear regression, we used
ordinary least squares. For the random forest baseline and XGBoost, the
hyperparameter grids can be seen below.

Random forest:
\begin{itemize}
    \item Number of estimators: [30, 100, 200] 
    \item Max depth: [3, 5, 10]
    \item Max features: [All, sqrt]
\end{itemize}

XGBoost grid:
\begin{itemize}
    \item Number of estimators: [30, 100, 200]
    \item Subsample: [0.6, 0.8, 1.0]
    \item Column sample by tree: [0.6, 0.8, 1.0]
    \item Max depth: [3, 5]
\end{itemize}

During fine-tuning, continuous features were normalized using the mean
and standard deviation of the current fold in K-fold cross-validation.

\newpage
\onecolumn
\section{Pre-training loss curves}\label{appendix_loss}

\begin{figure*}[htbp]
\centering
\includegraphics[width=\textwidth]{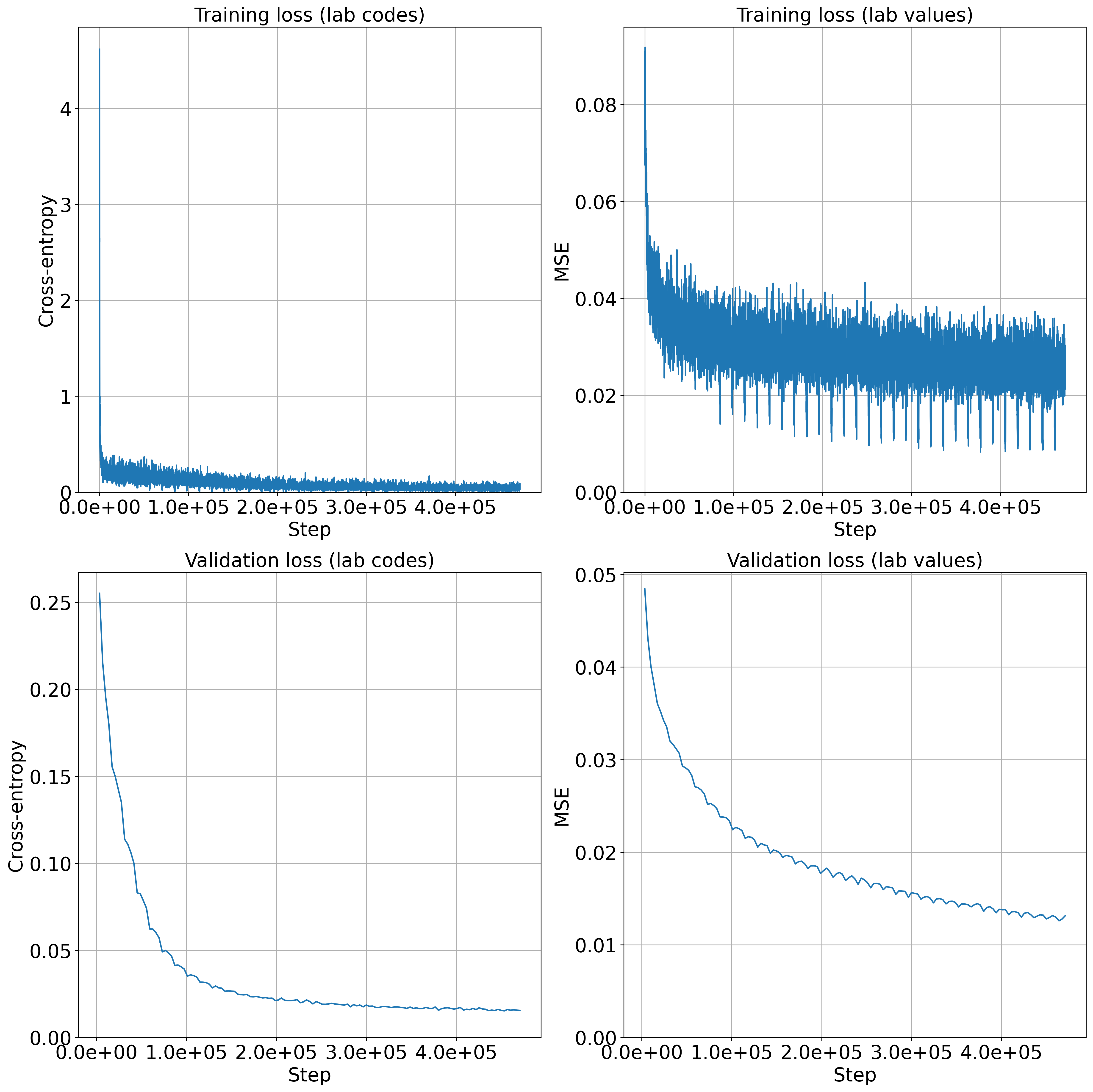}
\caption{Pre-training loss curves for \labrador, the continuous
Transformer. \textbf{Top:} The training loss for both
the lab code (cross-entropy) and lab value (MSE) prediction heads. \textbf{Bottom:} The validation cross-entropy and MSE.}
\label{fig_labrador_loss}
\end{figure*}

\clearpage

\begin{figure*}[htbp]
\centering
\includegraphics[width=\textwidth]{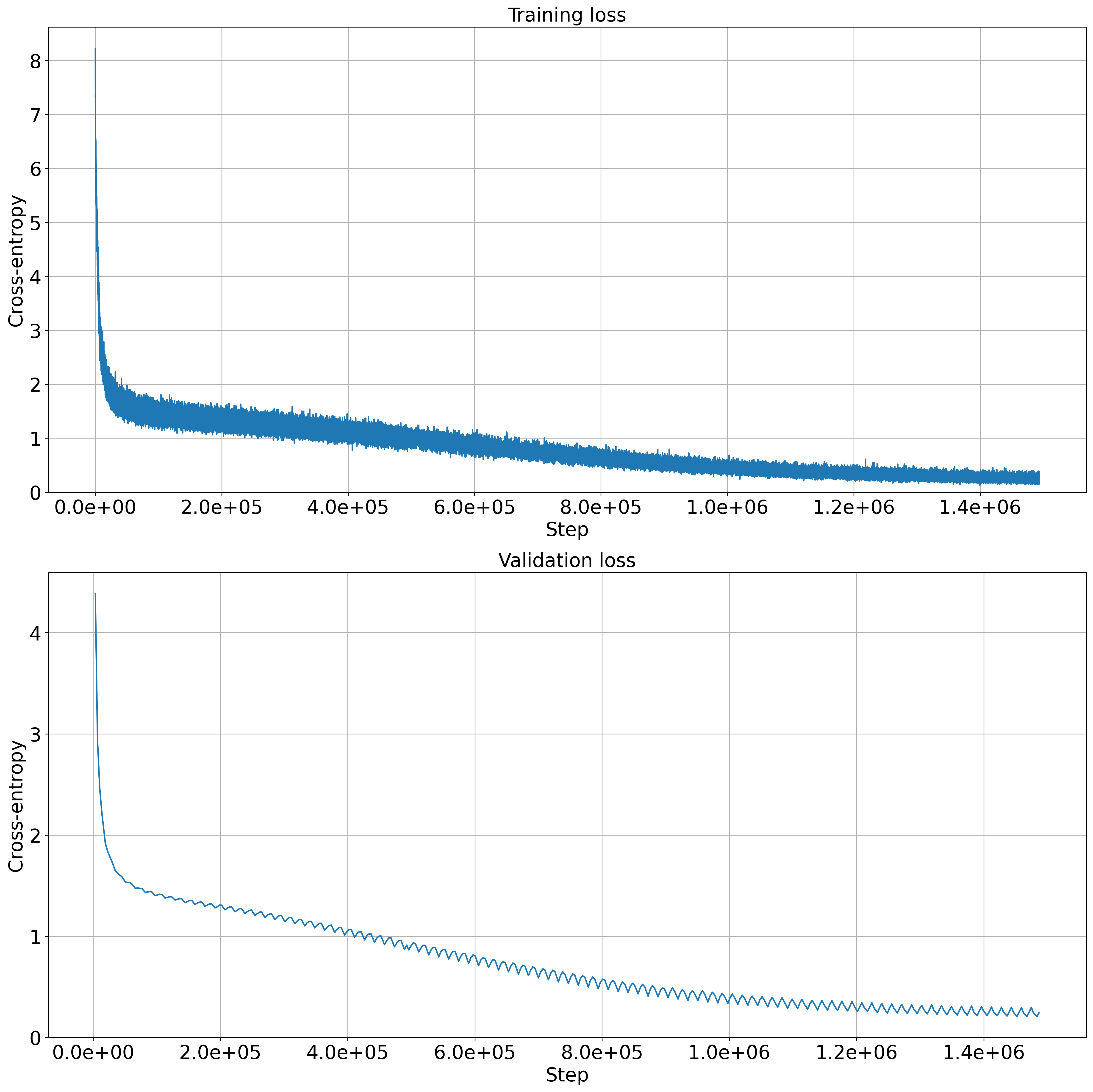}
\caption{Pre-training loss curves for BERT. \textbf{Top:} The training cross-entropy. \textbf{Bottom:} The validation cross-entropy.}
\label{fig_bert_loss}
\end{figure*}

\clearpage

\section{Lab value correlation plot}\label{appendix_corrmatrix}

\begin{figure*}[htbp]
\centering
\includegraphics[width=0.65\linewidth]{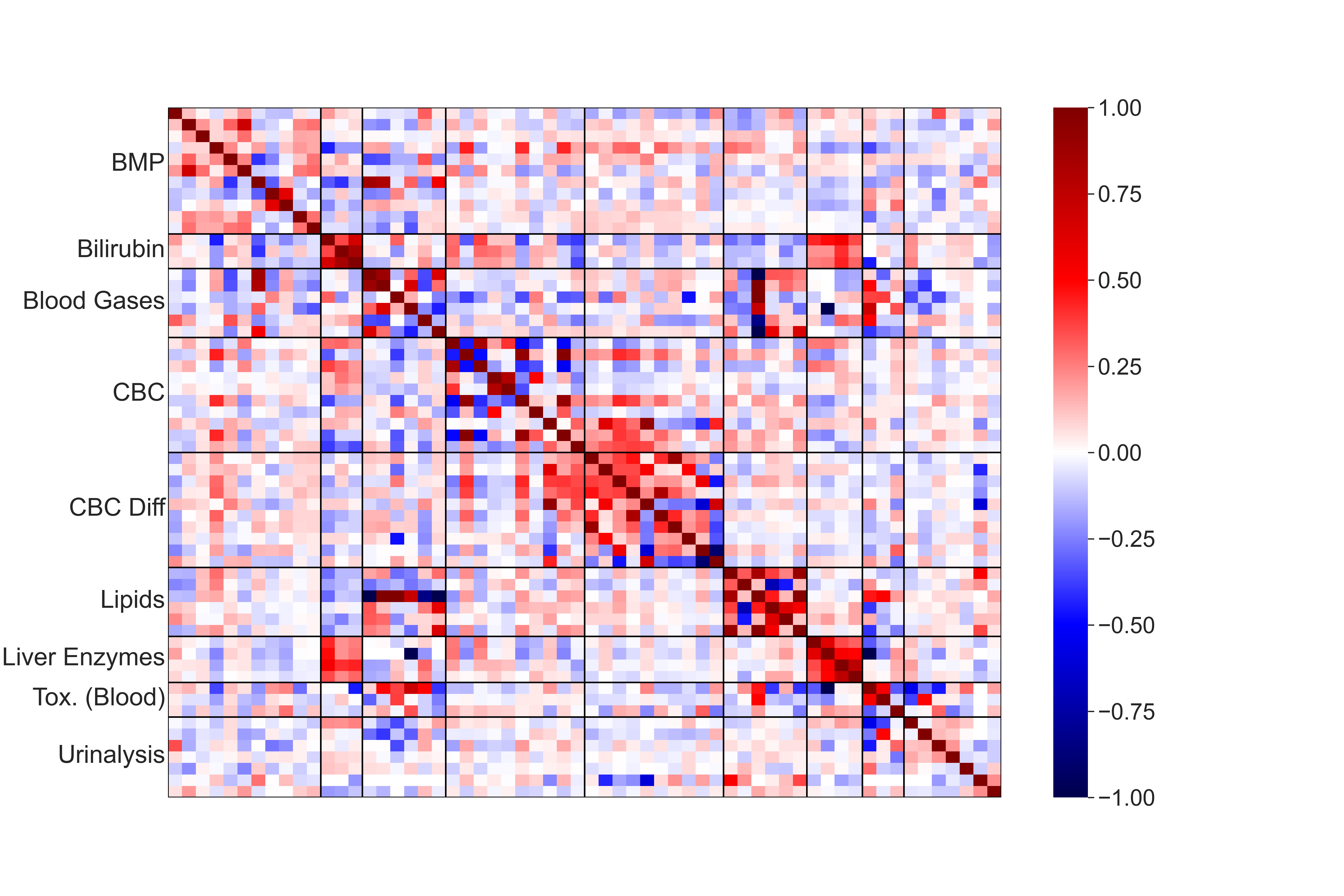}
\caption{Lab value correlation matrix. A 70 x 70 matrix of Pearson
correlation coefficients between each pair of the 70 most frequently
ordered lab tests in MIMIC-IV. The 70 tests
are grouped by the panel that they are most frequently ordered with
(see Appendix \ref{appendix_labpanels} for panel definitions). Black lines
divide the matrix into panel-by-panel correlation blocks.}
\label{fig_corrmatrix}
\end{figure*}

Figure \ref{fig_corrmatrix} is a heatmap of the Pearson correlation coefficients between each pair of the 70 most frequently ordered lab tests in MIMIC-IV. Labs are grouped by their panel membership and black lines separate these panels. For example, the top left box contains the correlation coefficients within the Basic
Metabolic Panel (BMP), whereas the box at row 6, column 3
represents the inter-panel correlation between lipid tests and blood gas tests. This allows us to compare intra-panel (main diagonal) to inter-panel correlation structure (off-diagonal).

\section{Lab panel definitions}\label{appendix_labpanels}
\begin{table*}[h]
\centering
\caption{Definition of the MIMIC-IV bilirubin panel.}\label{panel_bilirubin}
\small
\begin{tabular}{ l l l l l }
\toprule
\makecell{MIMIC-IV \\ ItemID} & \centering{Test Name} & \makecell{MIMIC-IV \\ Test Fluid} & \makecell{MIMIC-IV \\ Test Category} & LOINC Code \\
\midrule
50885 & Bilirubin, Total & Blood & Chemistry & 1975-2 \\
50883 & Bilirubin, Direct & Blood & Chemistry & 1968-7 \\
50884 & Bilirubin, Indirect & Blood & Chemistry & 1971-1 \\
\bottomrule
\end{tabular}
\end{table*}

\begin{table*}[h]
\centering
\caption{Definition of the MIMIC-IV blood gas panel. These tests do not have LOINC codes.}\label{panel_bloodgasses}
\small
\begin{tabular}{ l l l l}
\toprule
\makecell{MIMIC-IV \\ ItemID} & \centering{Test Name} & \makecell{MIMIC-IV \\ Test Fluid} & \makecell{MIMIC-IV \\ Test Category} \\
\midrule
50820 & pH & Blood & Blood Gas \\
50821 & pO2 & Blood & Blood Gas \\
50802 & Base Excess & Blood & Blood Gas \\
50804 & Calculated Total CO2 & Blood & Blood Gas \\
50818 & pCO2 & Blood & Blood Gas \\
50813 & Lactate & Blood & Blood Gas \\
\bottomrule
\end{tabular}
\end{table*}

\begin{table*}[h]
\centering
\caption{Definition of the MIMIC-IV basic metabolic panel (BMP).}\label{panel_bmp}
\small
\begin{tabular}{ l l l l l }
\toprule
\makecell{MIMIC-IV \\ ItemID} & \centering{Test Name} & \makecell{MIMIC-IV \\ Test Fluid} & \makecell{MIMIC-IV \\ Test Category} & LOINC Code \\
\midrule
50912 & Creatinine & Blood & Chemistry & 2160-0 \\
51006 & Urea Nitrogen & Blood & Chemistry & 3094-0 \\
50971 & Potassium & Blood & Chemistry & 2823-3 \\
50983 & Sodium & Blood & Chemistry & 2951-2 \\
50902 & Chloride & Blood & Chemistry & 2075-0 \\
50882 & Bicarbonate & Blood & Chemistry & 1963-8 \\
50868 & Anion Gap & Blood & Chemistry & 1863-0 \\
50931 & Glucose & Blood & Chemistry & 6777-7 \\
50893 & Calcium, Total & Blood & Chemistry & 2000-8 \\
50960 & Magnesium & Blood & Chemistry & 2601-3 \\
50970 & Phosphate & Blood & Chemistry & 2777-1 \\
\bottomrule
\end{tabular}
\end{table*}

\begin{table*}[h]
\centering
\caption{Definition of the MIMIC-IV complete blood count (CBC) panel.}\label{panel_cbc}
\small
\begin{tabular}{ l l l l l }
\toprule
\makecell{MIMIC-IV \\ ItemID} & \centering{Test Name} & \makecell{MIMIC-IV \\ Test Fluid} & \makecell{MIMIC-IV \\ Test Category} & LOINC Code \\
\midrule
51221 & Hematocrit & Blood & Hematology & 4544-3 \\
51265 & Platelet Count & Blood & Hematology & 777-3 \\
51222 & Hemoglobin & Blood & Hematology & 718-7 \\
51301 & White Blood Cells & Blood & Hematology & 804-5 \\
51249 & MCHC & Blood & Hematology & 786-4 \\
51279 & Red Blood Cells & Blood & Hematology & 789-8 \\
51250 & MCV & Blood & Hematology & 787-2 \\
51248 & MCH & Blood & Hematology & 785-6 \\
51277 & RDW & Blood & Hematology & 788-0 \\
52172 & RDW-SD & Blood & Hematology & N/A \\
\bottomrule
\end{tabular}
\end{table*}

\begin{table*}[h]
\centering
\caption{Definition of the MIMIC-IV complete blood count with differential panel.}\label{panel_cbc_diff}
\small
\begin{tabular}{ l l l l l }
\toprule
\makecell{MIMIC-IV \\ ItemID} & \centering{Test Name} & \makecell{MIMIC-IV \\ Test Fluid} & \makecell{MIMIC-IV \\ Test Category} & LOINC Code \\
\midrule
51256 & Neutrophils & Blood & Hematology & 761-7 \\
51244 & Lymphocytes & Blood & Hematology & 731-0 \\
51254 & Monocytes & Blood & Hematology & 742-7 \\
51146 & Basophils & Blood & Hematology & 704-7 \\
51200 & Eosinophils & Blood & Hematology & 711-2 \\
52075 & Absolute Neutrophil Count & Blood & Hematology & 751-8 \\
52073 & Absolute Eosinophil Count & Blood & Hematology & 711-2 \\
52074 & Absolute Monocyte Count & Blood & Hematology & 742-7 \\
51133 & Absolute Lymphocyte Count & Blood & Hematology & 731-0 \\
52069 & Absolute Basophil Count & Blood & Hematology & 704-7 \\
\bottomrule
\end{tabular}
\end{table*}

\begin{table*}[h]
\centering
\caption{Definition of the MIMIC-IV lipid panel.}\label{panel_lipid}
\small
\begin{tabular}{ l l l l l }
\toprule
\makecell{MIMIC-IV \\ ItemID} & \centering{Test Name} & \makecell{MIMIC-IV \\ Test Fluid} & \makecell{MIMIC-IV \\ Test Category} & LOINC Code \\
\midrule
51000 & Triglycerides & Blood & Chemistry & 1644-4 \\
50907 & Cholesterol, Total & Blood & Chemistry & 2093-3 \\
50904 & Cholesterol, HDL & Blood & Chemistry & 2086-7 \\
50903 & Cholesterol Ratio (Total/HDL) & Blood & Chemistry & 9322-9 \\
50905 & Cholesterol, LDL, Calculated & Blood & Chemistry & 2090-9 \\
50906 & Cholesterol, LDL, Measured & Blood & Chemistry & N/A \\
\bottomrule
\end{tabular}
\end{table*}

\begin{table*}[h]
\centering
\caption{Definition of the MIMIC-IV liver function test (LFT) panel.}\label{panel_lft}
\small
\begin{tabular}{ l l l l l }
\toprule
\makecell{MIMIC-IV \\ ItemID} & \centering{Test Name} & \makecell{MIMIC-IV \\ Test Fluid} & \makecell{MIMIC-IV \\ Test Category} & LOINC Code \\
\midrule
50861 & Alanine Aminotransferase (ALT) & Blood & Chemistry & 1742-6 \\
50878 & Asparate Aminotransferase (AST) & Blood & Chemistry & 1920-8 \\
50863 & Alkaline Phosphatase & Blood & Chemistry & 6768-6 \\
50927 & Gamma Glutamyltransferase & Blood & Chemistry & 2324-2 \\
\bottomrule
\end{tabular}
\end{table*}

\begin{table*}[h]
\centering
\caption{Definition of the MIMIC-IV toxicology (blood) panel.}\label{panel_tox_blood}
\small
\begin{tabular}{ l l l l l }
\toprule
\makecell{MIMIC-IV \\ ItemID} & \centering{Test Name} & \makecell{MIMIC-IV \\ Test Fluid} & \makecell{MIMIC-IV \\ Test Category} & LOINC Code \\
\midrule
50922 & Ethanol & Blood & Chemistry & 5642-4 \\
50856 & Acetaminophen & Blood & Chemistry & 3297-9 \\
50981 & Salicylate & Blood & Chemistry & 4023-8 \\
\bottomrule
\end{tabular}
\end{table*}

\begin{table*}[ht]
\centering
\caption{Definition of the MIMIC-IV toxicology (urine) panel.}\label{panel_tox_urine}
\small
\begin{tabular}{ l l l l l }
\toprule
\makecell{MIMIC-IV \\ ItemID} & \centering{Test Name} & \makecell{MIMIC-IV \\ Test Fluid} & \makecell{MIMIC-IV \\ Test Category} & LOINC Code \\
\midrule
51092 & Opiate Screen, Urine & Urine & Chemistry & N/A \\
51079 & Cocaine, Urine & Urine & Chemistry & N/A \\
51071 & Amphetamine Screen, Urine & Urine & Chemistry & N/A \\
51090 & Methadone, Urine & Urine & Chemistry & N/A \\
51074 & Barbiturate Screen, Urine & Urine & Chemistry & 3377-9 \\
51989 & Oxycodone & Urine & Chemistry & N/A \\
51089 & Marijuana & Urine & Chemistry & N/A \\
\bottomrule
\end{tabular}
\end{table*}

\begin{table*}[h]
\centering
\caption{Definition of the MIMIC-IV urinalysis panel.}\label{panel_urinalysis}
\small
\begin{tabular}{lllll}
\toprule
\makecell{MIMIC-IV \\ ItemID} & \centering{Test Name} & \makecell{MIMIC-IV \\ Test Fluid} & \makecell{MIMIC-IV \\ Test Category} & LOINC Code \\
\midrule
51506 & Urine Appearance & Urine & Hematology & 5767-9 \\
51498 & Specific Gravity & Urine & Hematology & 5811-5 \\
51491 & pH & Urine & Hematology & 5803-2 \\
51478 & Glucose & Urine & Hematology & 5792-7 \\
51492 & Protein & Urine & Hematology & 5804-0 \\
51484 & Ketone & Urine & Hematology & 5797-6 \\
51464 & Bilirubin & Urine & Hematology & 5770-3 \\
51487 & Nitrite & Urine & Hematology & 5802-4 \\
51514 & Urobilinogen & Urine & Hematology & 5818-0 \\
51087 & Length of Urine Collection & Urine & Chemistry & N/A \\
\bottomrule
\end{tabular}
\end{table*}